# Location of Single Neuron Memories in a Hebbian Network

Krishna Chaithanya Lingashetty

*Abstract-* **This paper reports the results of an experiment on the use of Kak's B-Matrix approach to spreading activity in a Hebbian neural network. Specifically, it concentrates on the memory retrieval from single neurons and compares the performance of the B-Matrix approach to that of the traditional approach.**

## I. Introduction

For many years, scientists have wondered as to how the human brain functions [1]-[10]. All that we know is that there is a complex network of neurons in the brain and these neurons learn by electrochemical signals which are generated by various organs of the body. Hence the need for the simulation and development of an artificial neural network is required for the better understanding of how a biological neural network might work.

One of the approaches taken for the construction of an artificial neural network is the feedback network with indexed memory retrieval. This particular method, developed by Kak, is called the B-matrix Approach [9],[10]. The B-matrix approach is a generator model for the neural network memory retrieval. By this, we mean that the activity starts from one neuron and then spreads to the adjacent neurons to increase the fragment length by one. The obtained fragment is then fed back to the network recursively until the entire memory is generated.

How the B-matrix approach works together with a specified proximity matrix for the neurons was recently shown by Kak [9]. In this paper, we perform experiments to see the relationship of single neuron memories to their location. The manner in which the location of memories scales up and the capacity for this storage have been estimated by performing experiments on a large number of networks with random memories.

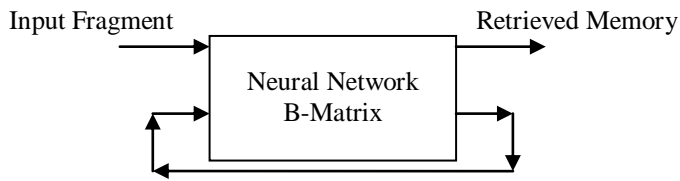

Figure 1. Obtaining a memory from a fragment

Krishna Chaithanya Lingashetty is with the Computer Science Department, Oklahoma State University, Stillwater, OK 74078 USA (phone: 817-235-2396; e-mail: krishna.lingashetty@okstate.edu).

## II  Preliminaries

### A. B-Matrix Approach

The neural network is trained by the Hebbian rules of learning. For various aspects related to storage of memories in neural networks, see [11]-[17]. References [18]-[21] deal with larger neuroscience issues concerning memories.

Hebbian learning assumes that the synaptic strength of neurons that fire together strengthens and that of those that don't gets weakened. The interconnection matrix T is calculated as, $T = \Sigma\, x^{(i)}x^{(i)t}$, where the memories are binary column vectors($x^i$), composed of {-1, 1} and the diagonal terms are taken to be zero. To verify if a particular memory $x^i$ is stored, the following condition should be valid,

$$x^i = sgn(T \cdot x^{(i)}) \qquad (1)$$

where *sgn()* is the Signum function, with the exception that at k = 0, sgn(k) = 1. Recollection of memories in the B-Matrix Approach is by using the lower triangular matrix *B*, constructed as,

$$T = B + B^t \qquad (2)$$

As shown in Figure 1, the activity starts from one single neuron and then spreads to additional neurons as determined by the Proximity Matrix which stores the geometric proximity between neurons. Starting with the fragment $f^{\,1}$, the updating proceeds as:

$$f^{\,i} = sgn(B \cdot f^{\,i-1}), \qquad (3)$$

where $f^{\,i}$ is the $i^{th}$ iteration of the generator model. Notice that the $i^{th}$ iteration of the generator model produces only the value of the $i^{th}$ binary index of the vector memory but does not alter the 'i-1' values already present.

### B. Proximity Martix and the Neural Network

The proximity matrix is a matrix which holds the measure of Geometric Proximity of each of the neurons with every other neuron in the network. The proximity matrix gives us an insight into the pattern in which a particular neural network retrieves memories through spreading activity.

The neural network of 'n' neurons can be thought of as a three dimensional network of 'n' nodes interconnected with each other with varying proximity between each node. We can construct a two dimensional graph of the network as a polygon of 'n' sides with all diagonals connected and each corner being a node. For example, consider the neural network of 6 nodes as shown in figure 2.

Let us assume without loss of generality that this network is already trained with a set of memories. When retrieving a memory from this network, the activity starts from one node and then spreads to the adjacent nodes as described by the proximity matrix and hence, retrieves the memory. Consider that the activity starts at the second neuron and spreads from there on. If the order given by the proximity matrix is [ 2 5 3 1 4 6 ], then memory retrieval proceeds as shown in the figure.

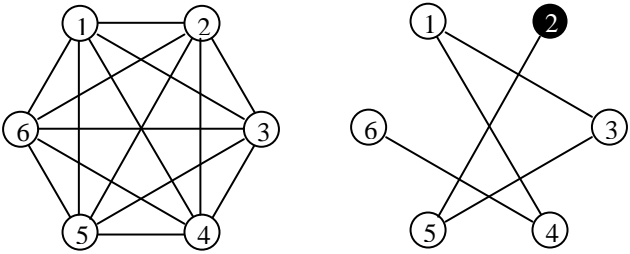

Figure 2. Graph showing the Neural Network and Activity Spread from Neuron 2.

Each neuron is capable of storing and retrieving a particular memory by the spread of activity as long as the network is not overwhelmed with information. By the above example, it is possible to retrieve the right memories if we do know what the index of the neuron to be stimulated is, and what should be used to stimulate that selected neuron. Hence indexing plays a major role in the retrieval of memories. To better understand how a complex neural network might function, we introduce the concept of a sub-neural net and an indexing neural net. A very different approach to indexing is provided in [13],[14].

### C. Sub-Neural Nets and Indexing Neural Nets

Consider a complex neural network, similar to the human brain, which can store and retrieve memories. This entire neural network can be further divided into much smaller, functional units of networks with the same capabilities as discussed above. In this Sub-Neural Net, each neuron is connected with every other neuron in the Sub-Neural Net and this network has a proximity matrix associated with it based on the proximity of the neurons.

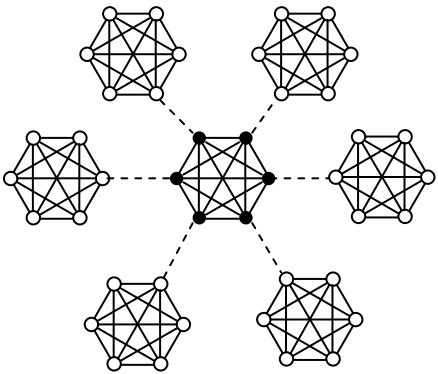

Figure 3. An Indexing Neural Net surrounded by Sub-Neural Nets

As mentioned above, it is possible to retrieve the right memories if we know which neuron to stimulate and what to stimulate this neuron with. Since a complex neural network is a very large, it is safe to assume that the network is made up of such small sub-neural networks, which are again interconnected with each other to retrieve and store memories. Hence, we can visualize this complex network as a network of networks, made up of components such as the sub-neural networks. This means, collectively, these sets of Sub-Neural Nets form a Neural Network that is in-turn managed by an Indexing Neural Network among the sub-neural networks, which may help in the retrieval of memories by clamping a certain set of neurons to retrieve information. A possible structure is illustrated in figure 3.

### III. EXPERIMENTAL SETUP

The proposed method was programmed in Java and was set to carry out the above procedure for a given number of iterations. In each iteration, the neural network is trained incrementally, one memory at a time, i.e., once a memory is fed into the neural network, the process of counting the number of stored memories(using (1)) and the number of retrieved memories(using (3)) is executed. After execution, the program generates the following output

1) A plot of the average of the number of memories stored against the number of memories fed to the network.
2) A plot of the average of the number of memories retrieved against the number of memories fed to the network.
3) A 2-D graph of the neural network with Highlighted Generators.

### A. Calculation of the Proximity Matrix

Since we do not have an already existing neural network and that we are simulating one, we also need to create the proximity matrix for the simulated network. As the labeling of neurons as 1,2,…n comes from the fact that we select a neuron to be labeled as 1 and the proximity order or the update order defines the labeling of the rest of the neurons as 2,3,4… in the increasing order of synaptic strength. Since the simulated neural network is already labeled, we need to create a proximity matrix such that it has the update order of [ 1 , 2 , 3 , …n]. Having established this, we define a bound on the maximum geometric proximity value possible (increasing order of proximity matrix values implies decrease in the synaptic strength of neurons). Consider that the maximum possible value for any element in the proximity matrix can never exceed (n-1). So the rest of the elements (excluding the update order of [ 1 , 2 , 3 , …n] and having all the diagonal elements to be zero) are filled with random values in between 0 and (n-1). This implementation works fine, except that we are not constructing a "fair" proximity matrix.

Since we are pre-specifying a proximity order of [ 1 , 2 , 3,…n ], it is quite possible that neuron-1 might have a very high geometric proximity with neuron-2 and hence, always tend to produce an update order that starts with [ 2 , 1 , …. ]. To give a fair chance for all the other neuron to be able to be geometrically closer to such neurons, the assignment of the proximity values for the update order [ 1 , 2 , ….n ] should be closer to the mean of all the possible values for the proximity matrix. Hence, the values to be associated with the update order of [ 1 , 2 , …n ] need to be closer to n/2.

## IV. RESULTS

### A. Relationship between the size of the network and the number of one-bit generators

As can be noted from the graphs, the number of retrieved memories does increase with the increase in the fed memories until a certain point and declines from there on. This is expected, as when we overwhelm the network with lots of information, the network becomes incapable of storing additional memories and may even lose the memories that it had already stored in it.

Looking at the graphs, we can notice that for smaller neural networks, the B-Matrix approach gives us a better chance at retrieving more memories as compared to the traditional approach. But as the number of memories increases, the memories retrieved by the B-Matrix approach fall below the traditional approach. Also as the size of the neural network increases, the effectiveness of the B-Matrix approach decreases. This can be observed in the 512 and the 1024 neural network example, as, the retrieved memories using the B-matrix approach never exceeds the traditional method.

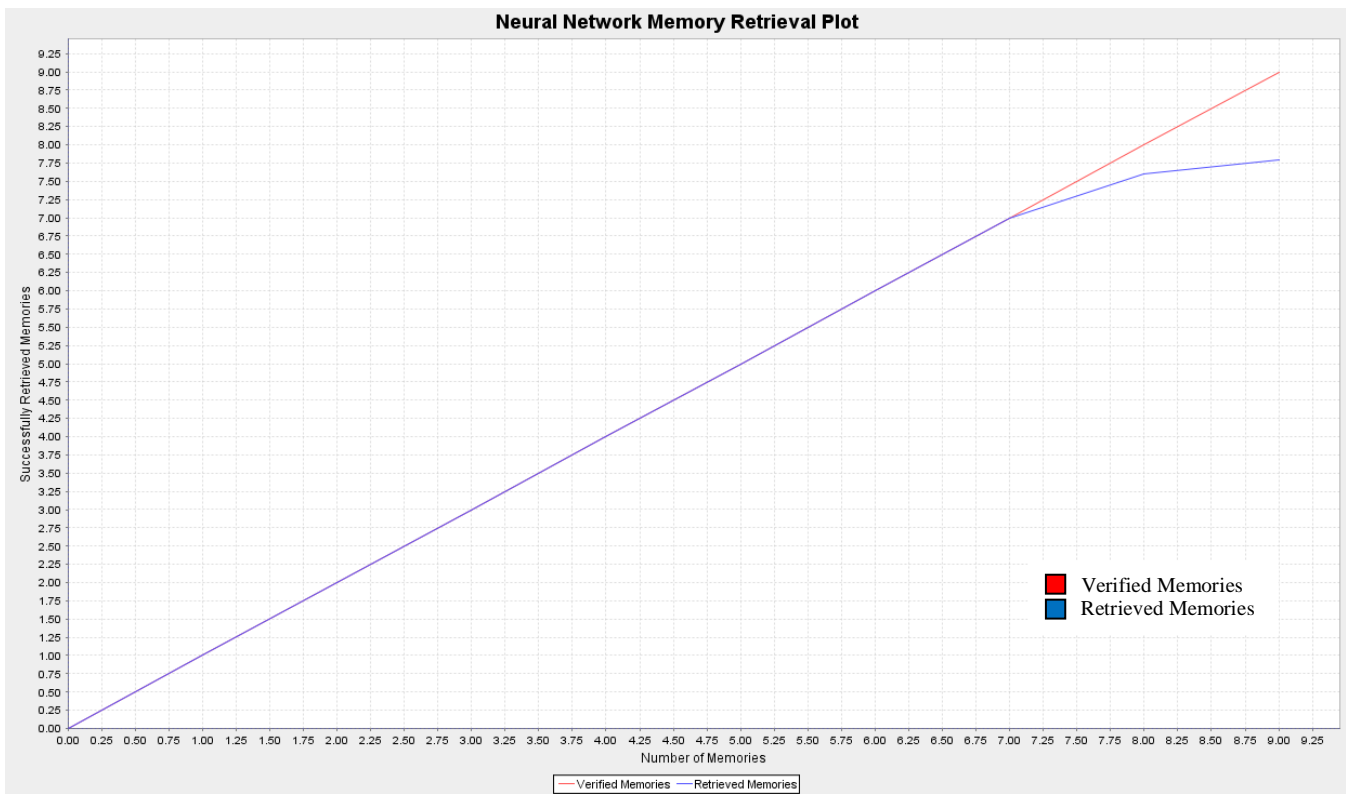

Figure 4. 1024 Neurons and 10 Iterations

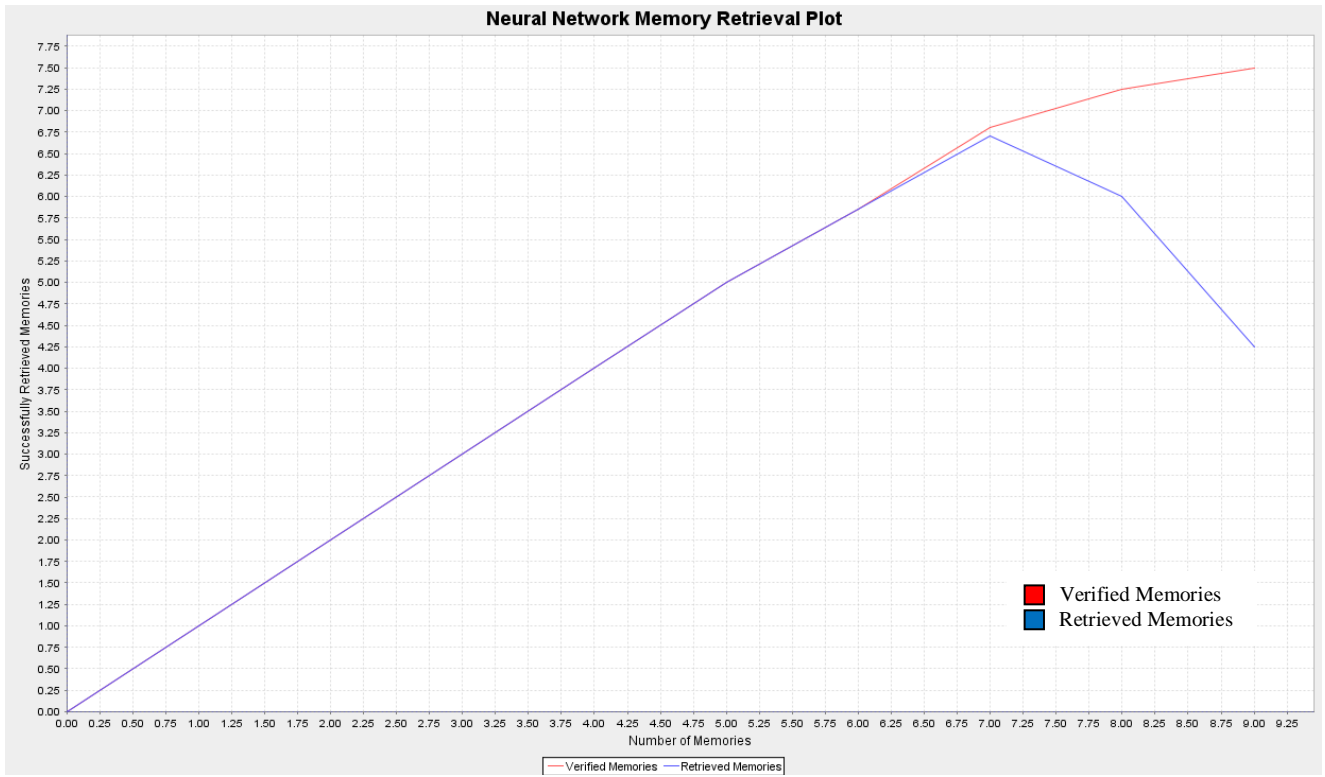

Figure 5. 512 Neurons and 100 Iterations

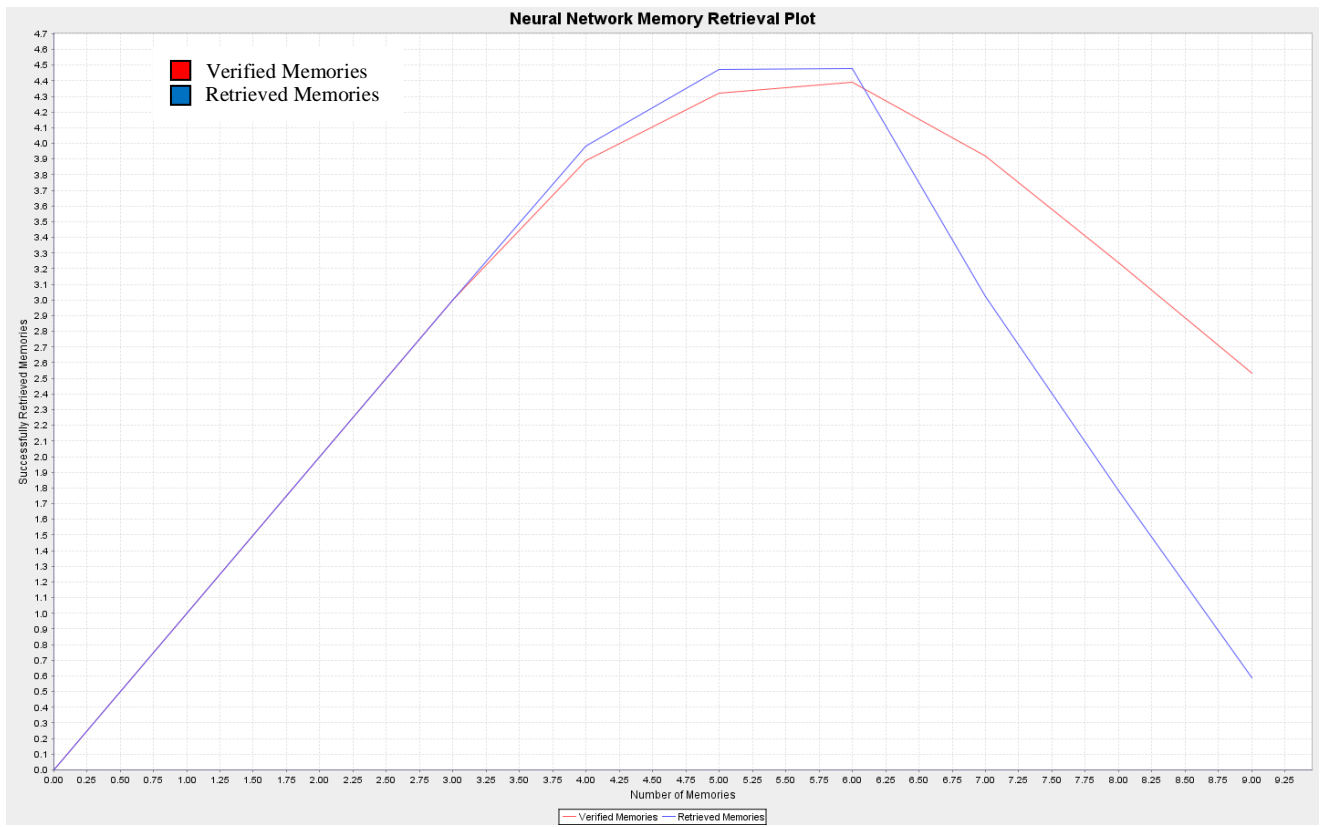

Figure 6. 256 Neurons and 100 Iterations

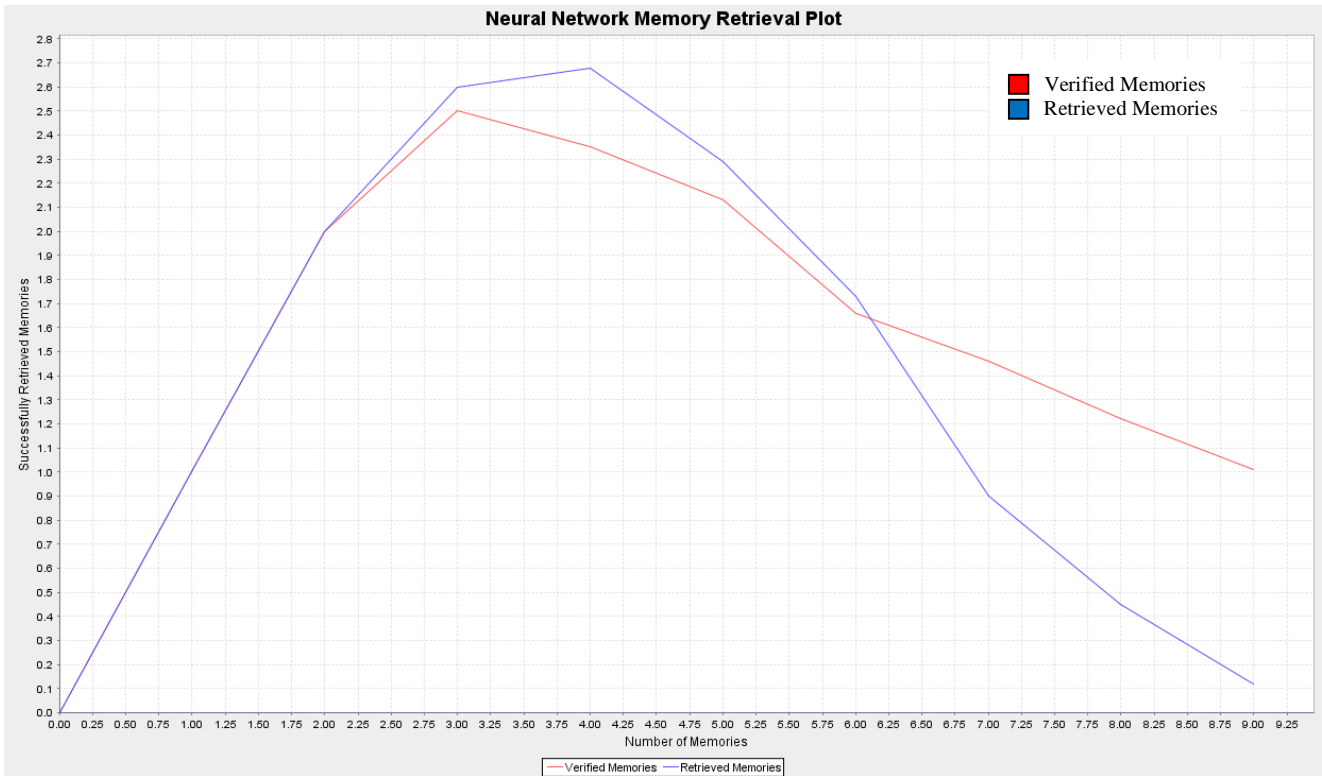

Figure 7. 128 Neurons and 100 Iterations

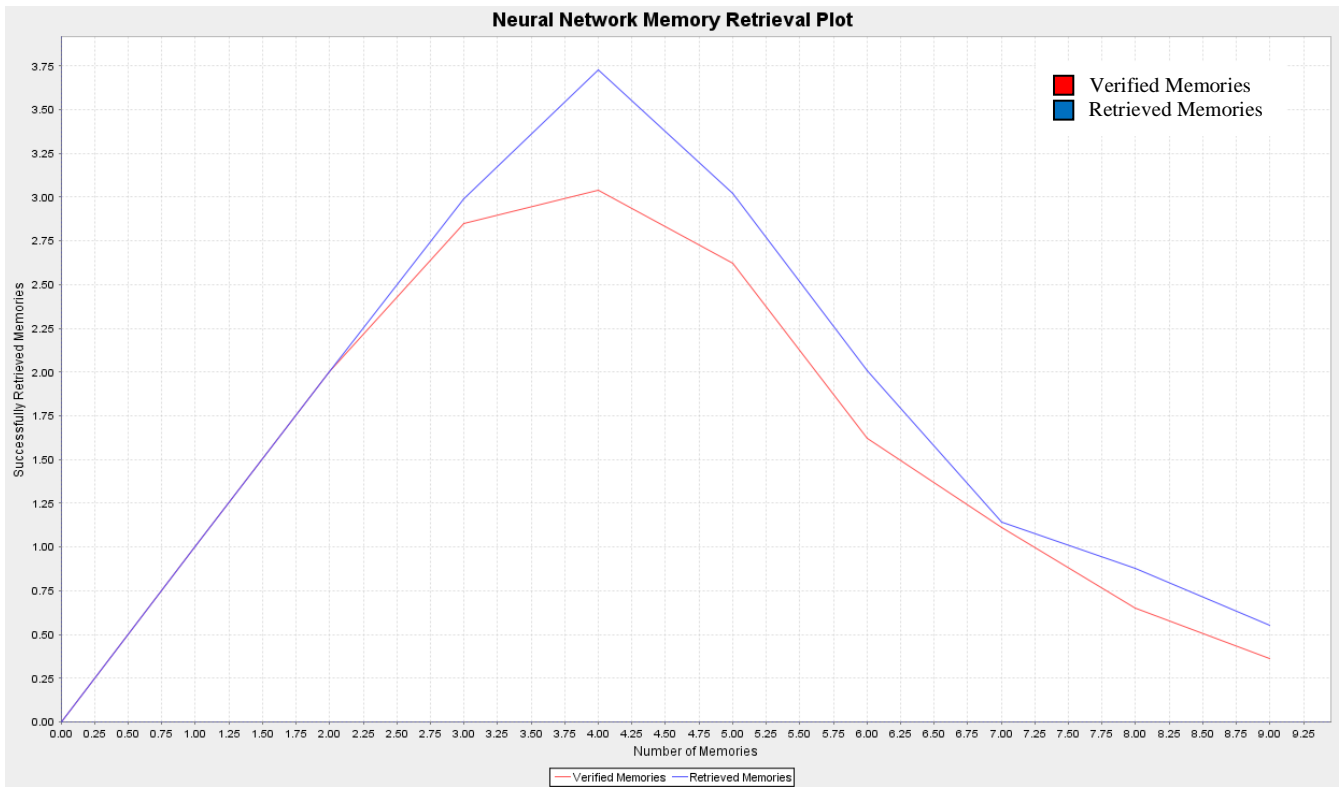

Figure 8. 64 Neurons and 100 Iterations

## B. The number of unique generators associated with a network

The following figures show the graphs of a Neural Network of 16, 32 and 64 neurons, trained with 4, 4 and 5 memories respectively. The highlighted neurons are the ones that have generated a memory successfully. The rest of the neurons are the ones that have not successfully generated any memory. Each color of the nodes corresponds to a different memory.

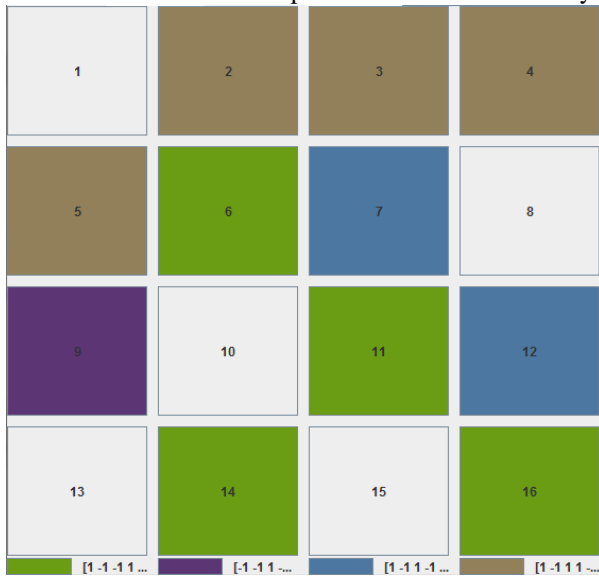

Figure 9. 2-D graph of Generators in a 16-Neural Network

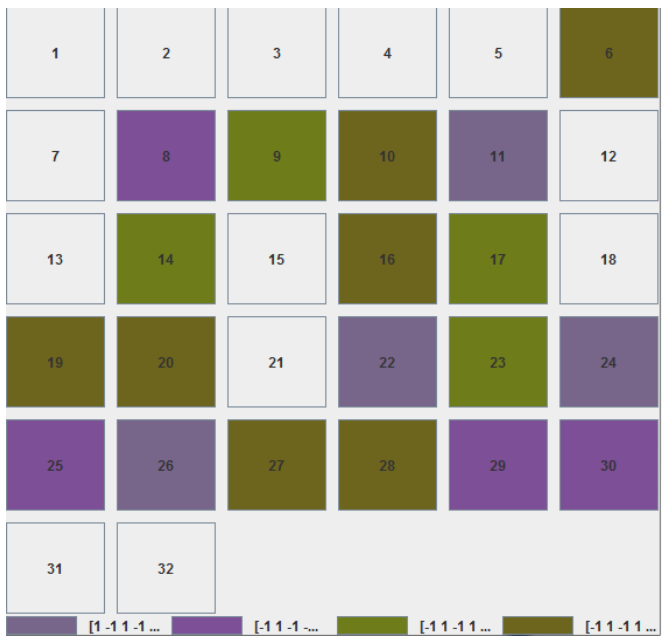

Figure 10. 2-D graph of Generators in a 32-Neural Network

From the given two graphs, we can see that as the size of the neural network grows, the number of neurons that can successfully generate a stored memory also increases, but the percentage of neurons that do not generate any memory increases (31.25%, 40.62%, and 75%). By this we can conclude that as the size of the network increases, the increase in the number of generators (memory retrieving single neurons) is not proportional. Another interesting aspect of consideration would be to notice the number of neurons generating a single memory. Even though the number of generators that produce a particular memory may increase with the increase in the size of the network, the percentage of generators that produce a particular memory is decreasing (25%, 21.87% and 7.81% in the best cases).

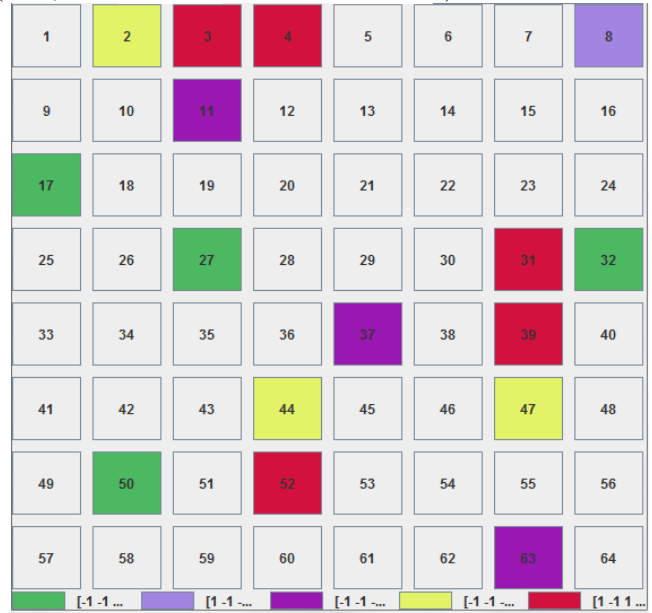

Figure 11. 2-D graph of Generators in a 64-Neural Network

## V. DISCUSSION

In the examination of indexing of memories, we have found out that triggering the right neurons with the right stimulus gives us the retrieved memory. The same may be attributed to the thinking or recollecting action performed by humans. When we are thinking about a past memory or an event, it is quite possible that we might stimulate an indexing neural network, which then stimulates its corresponding sub-neural networks until the pieces of memory are retrieved. If the retrieved fragments collectively are not the desired memory, then the indexing neural network may be subjected to another stimulus in anticipation of successful retrieval.

Hence further study needs to be conducted on the retrieval patterns and behavior of single generators and multi generators consisting of multiple neurons. Also, as this model is so heavily based upon the geometrical proximity of the neurons in the network, that there needs to be a more sophisticated or heuristic approach to the construction of the proximity matrix. Also, is the proximity matrix the only factor in deciding the update order? Or are there other ways in which the update order may be modified? If so, what is the complexity of such an approach? Several properties of such a network are still to be investigated.

ACKNOWLEDGMENT

This research was made possible by a grant from Center of Telecommunications and Network Security (CTANS).

REFERENCES

[1] S. Zeki, A Vision of the Brain, Blackwell Scientific Publications, Oxford, 1993.
[2] D.O. Hebb, *The Organization of Behavior*. Wiley, 1949.
[3] S. Kak, The three languages of the brain: quantum, reorganizational, and associative. In: K. Pribram, J. King (Eds.), Learning as Self-Organization, Lawrence Erlbaum, London, 1996, pp. 185-219.
[4] S. Kak, The honey bee dance language controversy. Mankind Quarterly 31, 357-365, 1991.
[5] M.F. Bear, B. Connors, and M. Paradiso, Neuroscience: Exploring the Brain. Lippincott Williams and Wilkins, 2006.
[6] S. Kak, Better web searches and prediction with instantaneously trained neural networks, IEEE Intell. Syst. 14 (6), 78–81, 1999.
[7] S. Kak, Can we define levels of artificial intelligence? *Journal of Intelligent Systems,* vol. 6, pp. 133-144, 1996.
[8] J.J. Hopfield, Neural networks and physical systems with emergent collective computational properties. *Proc. Nat. Acad. Sci. (USA)*, vol. 79, pp. 2554-2558, 1982.
[9] S. Kak, Single neuron memories and the network's proximity matrix. 2009. arXiv:0906.0798
[10] S. Kak, Feedback neural networks: new characteristics and a generalization. *Circuits, Systems, and Signal Processing*, vol. 12, pp. 263-278, 1993.
[11] S. Kak, Self-indexing of neural memories. *Physics Letters A,* vol. 143, pp. 293-296, 1990.
[12] S. Kak, State generators and complex neural memories. *Pramana*, vol. 38, pp. 271-278, 1992.
[13] S. Kak and M.C. Stinson, A bicameral neural network where information can be indexed. *Electronics Letters*, vol. 25, pp. 203-205, 1989.
[14] M.C. Stinson and S. Kak, Bicameral neural computing. *Lecture Notes in Computing and Control*, vol. 130, pp. 85-96, 1989.
[15] D.L. Prados and S. Kak, Neural network capacity using the delta rule. *Electronics Letters,* vol. 25, pp. 197-199, 1989.
[16] D. Prados and S. Kak, Non-binary neural networks. *Lecture Notes in Computing and Control,* vol. 130, pp. 97-104, 1989.
[17] S. Kak and J. F. Pastor, Neural networks and methods for training neural networks. *US Patent* 5,426,721, 1995.
[18] R.Q. Quiroga, L. Reddy, G. Kreiman, C. Koch, and I. Fried, Invariant visual representation by single neurons in the human brain. *Nature* **435**, pp. 1102–1107, 2005.
[19] W.M. Kistler and W. Gerstner, Stable Propagation of Activity Pulses in Populations of Spiking Neurons. Neural Computation, vol. 14, pp. 987-997, 2002.
[20] K.H. Pribram and J. L. King (eds.), *Learning as Self-Organization*. Mahwah, N. J.: L. Erlbaum Associates, 1996.
[21] K Grill-Spector, R. Henson, and A. Martin, Repetition and the brain: neural models of stimulus-specific effects. Trends in Cognitive Sciences, 2006.